\pdfoutput=1

\documentclass[11pt]{article}

\usepackage{ACL2023}

\usepackage{times}
\usepackage{latexsym}
\usepackage{graphicx}
\usepackage{booktabs}
\usepackage[T1]{fontenc}

\usepackage[utf8]{inputenc}

\usepackage{microtype}

\usepackage{inconsolata}

\title{LongSkywork: A Training Recipe for Efficiently Extending Context Length in Large Language Models}

\author{ {Liang Zhao, Tianwen Wei, Liang Zeng, Cheng Cheng, Liu Yang} \\
 {Peng Cheng, Lijie Wang, Chenxia Li, Xuejie Wu, Bo Zhu, Yimeng Gan} \\
 {Rui Hu, Shuicheng Yan, Han Fang, Yahui Zhou}
\thanks{\quad Email: \texttt{\{forename\}.\{surname\}@kunlun-inc.com}} \\
\quad \\
Skywork Team, Kunlun Inc.
}

\begin{document}
\maketitle
\begin{abstract}
We introduce LongSkywork, a long-context Large Language Model (LLM) capable of processing up to 200,000 tokens. We provide a training recipe for efficiently extending context length of LLMs. 
We identify that the critical element in enhancing long-context processing capability is to incorporate a long-context SFT stage following the standard SFT stage. A mere 200 iterations can convert the standard SFT model into a long-context model. To reduce the effort in collecting and annotating data for long-context language modeling, we develop two novel methods for creating synthetic data. These methods are applied during the continual pretraining phase as well as the Supervised Fine-Tuning (SFT) phase, greatly enhancing the training efficiency of our long-context LLMs. Our findings suggest that synthetic long-context SFT data can surpass the performance of data curated by humans to some extent.
LongSkywork achieves outstanding performance on a variety of long-context benchmarks. In the Needle test, a benchmark for long-context information retrieval, our models achieved perfect accuracy across multiple context spans. 
Moreover, in realistic application scenarios, LongSkywork-13B demonstrates performance on par with Claude2.1, the leading long-context model, underscoring the effectiveness of our proposed methods.

\end{abstract}

\section{Introduction}

The advancement of large language models~\cite{touvron2023llama1, anil2023palm, bai2023qwen, du2021glm, wei2023skywork}, exemplified by ChatGPT~\cite{gpt4_tech_report}, is greatly impacting the world. These models are being used in various fields, such as chitchat~\cite{ouyang2022training}, composition~\cite{fyfe2023cheat}, and dynamic agents~\cite{xi2023rise, bozkurt2023generative}. A key factor driving this progress is the ability of these models to handle inputs with extensive context. However, a notable limitation remains: because of limited resources, many existing language models are mainly trained on shorter texts.  For example, Llama1~\cite{touvron2023llama1} is trained on 2K context windows, while Llama2~\cite{touvron2023llama2} is trained on 4K context windows. Consequently, their effectiveness might be reduced when faced with longer-form prompts that are frequently encountered in real-world applications.


\begin{figure}[t]
    \centering
    \includegraphics[width=.48\textwidth]{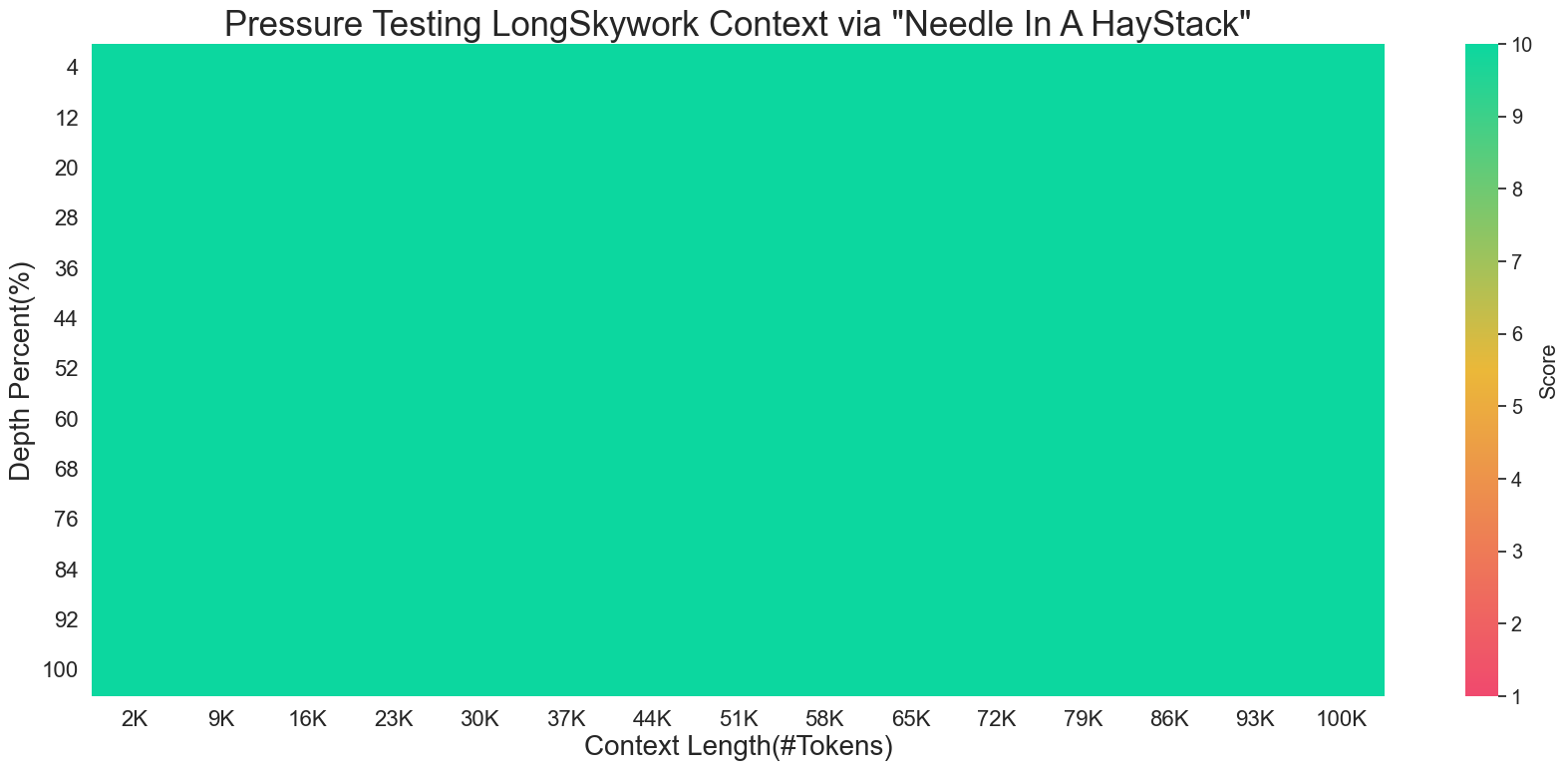}
    \caption{Needle Test~\cite{needletest} result for LongSkywork-13B. The Needle Test evaluates the key information retrieval capability of an LLM. }
    \label{fig:needle}
\end{figure}

The demand for LLMs that can understand and generate lengthy sequences efficiently has grown significantly, given their widespread use in applications that involve complex scenarios~\cite{liu2023lost, beltagy2020longformer, milne2020effectiveness}. Consequently, researchers have dedicated substantial efforts to enhance the Transformer architecture with long-context processing capability~\cite{roziere2023code}. This includes techniques such as continual pretraining with long data~\cite{xiong2023effective}, positional embedding interpolation~\cite{chen2023extending, peng2023yarn}, memory-efficient attention such as Flash attention~\cite{dao2022flashattention}. However, the most powerful long-context models currently available are proprietary systems, such as GPT4~\cite{gpt4_tech_report}, Claude~\footnote{https://www.anthropic.com/news/claude-2-1} and Moonshot~\footnote{https://kimi.moonshot.cn/}.
There remains a need for more information on how to develop high-performing, long-context language models for public use. Most open-sourced models that can process over 100k tokens are either base models unsuitable for assisting real-world tasks, or fine-tuned only on long dialogues~\cite{zeng2022glm}, exhibiting poor comprehension of lengthy contexts. Moreover, high-quality, natural long-form datasets are scarce for pretraining and supervised fine-tuning(SFT), presenting challenges in training powerful long-context LLMs.


To bridge this gap, we introduce LongSkywork, a long-context LLM capable of processing up to 200,000-token context windows, which is specifically tailored to facilitate in-depth comprehension and problem-solving across diverse scenarios. Furthermore, we offer critical insights into the developmental framework of LongSkywork.
Our findings suggest that the model's ability to handle extended contexts can be systematically augmented through a four-stage process. Beyond the conventional stages of pretraining and supervised fine-tuning, our approach incorporates specialized stages for long-context pretraining and long-context supervised fine-tuning. These additional stages are meticulously designed to bolster the model's proficiency in managing extended textual segments.
Significant performance enhancements during both long-context training phases were observed upon the integration of synthetically generated data. For the long-context pretraining phase, we propose a novel method where documents are splited into segments, which are then arranged in an interleaved fashion to form pseudo long-context pretraining samples, a technique we term ``Chunk Interleaved Pretraining (CIP)''. This technique obliges the model to process and integrate information over considerable textual distances.

During the long-context supervised fine-tuning phase, we present a method for generating queries and answers from tables. These tables are automatically synthesized by the program and referred to as ``Synthetic Lengthy Tables'' (SynL). This method mandates that the model not only extract relevant information from extensive tabular data but also engage in complex reasoning tasks under multiple constraints.
Additionally, we engineer specific tasks to enhance the model's global reasoning capabilities, such as the transformation and sorting of tables. These exercises are intricately designed to compel the model to develop a more comprehensive and nuanced understanding of intricate data structures and their inherent relationships.

Our extensive experiments indicate that the long-context SFT stage is crucial in enhancing the model's long context ability. Our approach is efficient, necessitating only hundreds of long-context pretraining and SFT iterations to furnish a short-text LLM with long context capability. LongSkywork performs well on the Needle in a Haystack test, as shown in the Figure~\ref{fig:needle}. 
Furthermore, we validate our model using both the long-context benchmark InfiniteBench~\cite{zhang2023infinitebench} and a real-world evaluation collected from online long-context question-answering. The results show that LongSkywork-13B has strong long-context retrieval capabilities, outperforming GPT4-128K and Claude2.1-200K in three retrieval-based tasks and achieving average results on par with Moonshot in InfiniteBench.
In real-world evaluations, LongSkywork-13B performs comparably to Claude2.1, but with significantly fewer parameters.

The contributions of this paper are summarized as follows:

\begin{itemize}
\item We introduce LongSkywork, a long-context LLM with up to 200K context windows. We provide a training recipe for efficiently extending context length in LLMs. 
\item We propose methods to construct synthetic data to accelerate the learning of long context for solving the data scarcity problem on training long-context LLMs. 
\item Our models show powerful capability in tasks that require long context retrieval and comprehensive ability. Our model is on par with GPT-4-128K and Claude2.1 in InfiniteBench. 
\end{itemize}

\begin{figure*}[t]
	\centering
 	
    \includegraphics[width=0.98\linewidth]{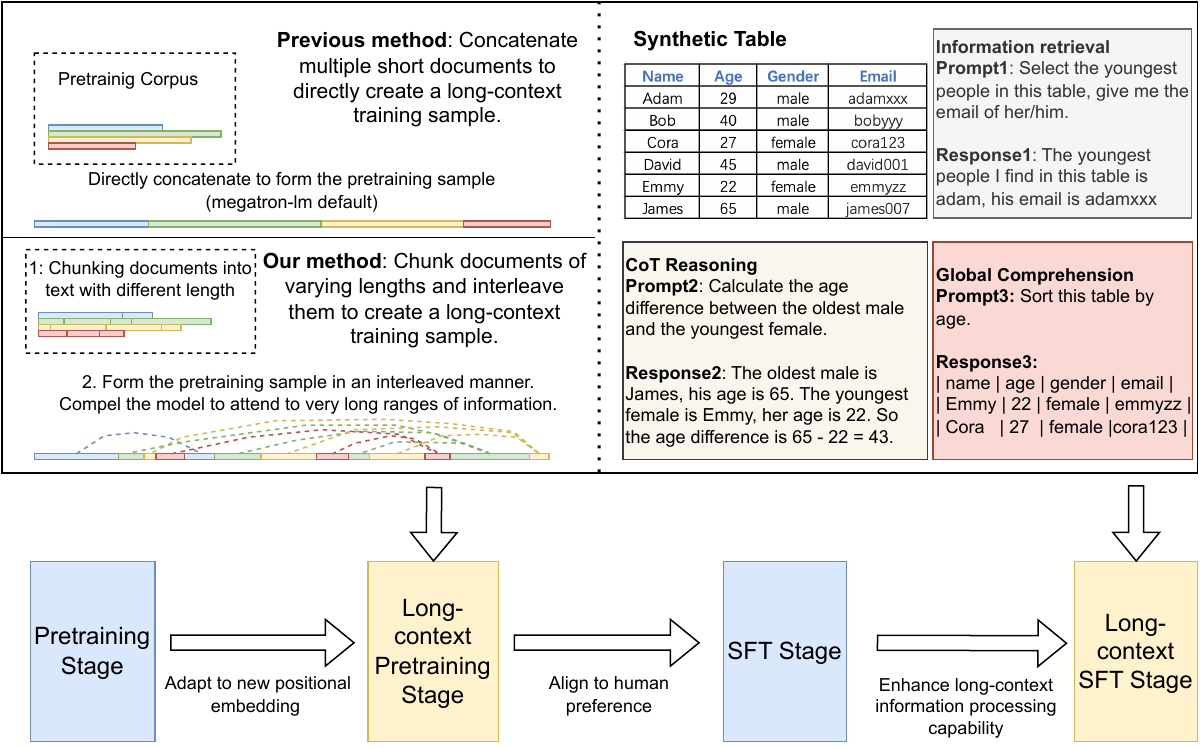}
	\caption{An illustration of our proposed methods. Except regular pretaining stage and SFT stage, we add two long-context related stages to empower LLMs with capability of long-context information processing.}\label{fig:model-details}
\end{figure*}

\section{Related work}
\subsection{Long-context LLMs}
Adapting Transformers~\cite{vaswani2017attention,dai2019transformer} and LLMs~\cite{touvron2023llama1} to handle long contexts is an important area of research. There are two lines of approaches related to our methods. The first one is the extended Rotary Position Embedding(RoPE). RoPE, proposed by ~\citet{su2024roformer}, is widely used in popular LLMs, such as Llama~\cite{touvron2023llama1,touvron2023llama2} and PaLM~\cite{anil2023palm}. It uses absolute positional embedding to represent relative position and provides a clear theoretical explanation in a complex space. Recently, many researchers found that RoPE positional embedding can effectively extend the inference context length with minimal or no finetuning~\cite{chen2023extending, peng2023yarn}. \citet{chen2023extending} proposed Positional Interpolation (PI), which applies linear scaling on each positional index from n to n/k, densifying the representation space to extend the farthest length by k times. However, simple linear scaling may result in limited space, which can impair the model's understanding of fine-grained distances. Based on Neural Tangent Kernel theory~\cite{jacot2018neural}, \citet{ntk_aware_rope} proposed to extend the extrapolated length by adjusting the base in the position function. These methods are referred to as NTK-aware Scaling RoPE (NTK-aware RoPE), which combines high-frequency extrapolation and low-frequency interpolation. Due to its simplicity and efficiency, many long-context LLMs exploit this approach for extrapolation~\cite{roziere2023code, xiong2023effective}.  Yarn~\cite{peng2023yarn} combines NTK-aware RoPE and positional interpolation. This method suggests not interpolating the higher frequency dimensions, while always interpolating the lower ones. 

The other line of related research involves study the training data in extending  context windows of LLMs. PI and YARN models are finetuned on the PG19 dataset~\cite{rae2019compressive}, which is a book dataset with significantly longer context. However, training on a different distribution than the pretraining stage may have a negative impact on the model's performance. \citet{xiong2023effective} suggested combining multiple short contexts to achieve the desired length, which also makes training more efficient. They also found that increasing the proportion of long text does not necessarily improve the model's performance on long context. In this paper, we show that by merely dividing the text into chunks and interleaving them to form a long-context pretraining sample, we can improve the model's ability to handle long context without altering the training distribution of the pretraining phase. Furthermore, previous works mainly focus on the long-context pretraining stage and provides limited insights into how the long-context SFT stage impacts the model's performance. 

\subsection{Synthetic data used in LLMs}

Training LLMs on human-collected data remains the predominant approach. However, human-generated data is limited and hard to collect, especially in long-context language modeling scenarios~\cite{xiong2023effective}. Besides, human-generated data will be exhausted in the next few years~\cite{benaich2020state}. Therefore, exploring the use of synthetic data to assist in LLM training and alignment is a promising research area. ~\citet{cao2023unnatural} found that GPT4 can almost perfectly reconstruct the original sentences from scrambled ones. As a result, researchers in the community speculate that GPT4 may utilize synthetic data for learning. Additionally, there are ample studies of using synthetic data during the SFT stage. \citet{wei2023simple} discovered that incorporating synthetic math data during SFT significantly outperforms finetuning using only human data. ~\citet{shao2023synthetic} suggested using a few handcrafted examples to prompt the model to generate more examples on its own and select effective demonstrations to elicit better reasoning.~\citet{sun2023principle} proposed to prompting LLMs to generate alignment SFT data based on several seed examples. However, those methods only focuses on the adoption of LLMs to generate synthetic data. The use of programming to generate synthetic data to assist in the SFT stage has not been extensively researched. In this paper, we use synthetic data in both pretraining and SFT stages, and shows that it can significantly boost the power of LLMs in long-context retrieval and understanding. 

\section{Methodology}
\subsection{Preliminaries}
Traditionally, training an LLM can be split into two stages. The first stage is referred to as the pretraining stage. It uses an auto-regressive language objective to generate a sequence $\textbf{y}=\{y_1, y_2, y_3, … y_n \}$ based on the previously generated tokens. Assuming that the language model is parameterized by $\theta$, the auto-regressive process can be described as follows:

$$
p_\theta(\textbf{y}) = p_{\theta}(y_t, y_{<t}).
$$

The pretraining stage is seen as injecting most of the knowledge into LLMs~\cite{zhou2023lima}. The second stage is referred to as the supervised finetuning stage, which can be seen as teaching the model to align format. An SFT training sample consists of a query, denoted as $\textbf{x}$, which is used as input, and an output $\textbf{y}$ that the model wants to generate conditioned on that prompt. Therefore, the training process in the SFT stage can be formalized as:
$$
p_\theta(\textbf{y}|\textbf{x}) = p_{\theta}(y_t, y_{<t}| \textbf{x}),
$$

where $\textbf{x}$ is masked and only $\textbf{y}$ produces gradient to update the parameters of the model.

\begin{figure}[t]
    \centering
    \includegraphics[width=.5\textwidth]{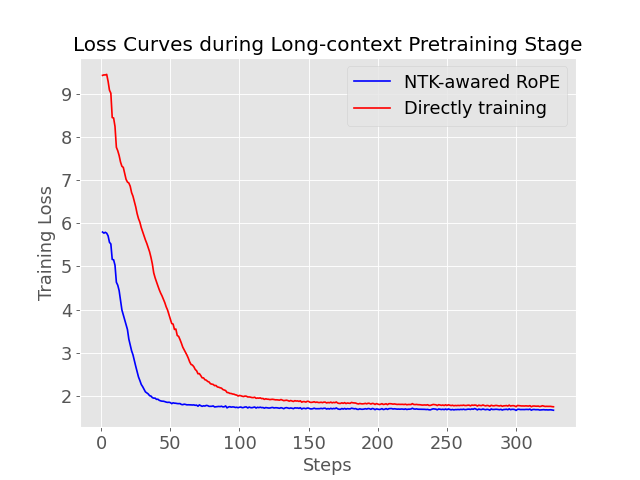}
    \caption{Loss curves during long-context pretraining stage. The training loss converges in just a few hundred iterations. NTK-aware RoPE converges faster than directly training.}
    \label{fig:long_context_pretrainig}
\end{figure}

Previous work~\cite{xiong2023effective} adds a long-context pretraining stage after the traditional pretraining stage. 
The long-context pretraining stage is designed to help the model adjust to the new context, mainly the new parameter of RoPE positional embedding.  As shown in Figure~\ref{fig:long_context_pretrainig}, only hundreds of long-context training iterations make the training loss converge. PI and YARN use book data to conduct long-context training. We empirically find that it makes the long-context pretraining stage learn a different distribution than that in pretraining stage. ~\citet{xiong2023effective} demonstrated that natural long context, such as a book or code repository, isn't necessary during the long-context training stage. Instead, we can concatenate short contexts to form a longer one. Surprisingly, this method can outperform a dataset with a higher long text ratio. Following~\citet{xiong2023effective}, we use the training data distribution from the pretraining stage and concatenate it to the long context to form the training data for the long context pretraining stage.

After conducting a standard SFT stage, we found that implementing a long-context SFT stage significantly improves retrieval and comprehension of long-context information. ~\citet{xiong2023effective} suggested merging normal SFT samples with long-context SFT samples and training them together.However, we noticed that when blending normal SFT with long-context SFT data, the gradient of the short context tends to overshadow the long-context SFT data and the model performs poorly in long-context tasks. This happens because the SFT stage only calculates the gradient for the response.
As seen in Table~\ref{tab:sft_data_distribution}, long-context SFT samples usually have a shorter response relative to their prompts. These prompts often include a context from Wikipedia or a book, followed by a related question. The answer to the question, typically brief, is often the key information in this context. Conversely, normal SFT data often involves creatively written queries with relatively longer responses to their prompts. 
Moreover, training a model with 13B parameters in a 4K context window yields a throughput of approximately 1775 tokens per second per GPU. However, when training with a 100K context window, the speed drops to about 180 tokens per second per GPU. If we combine short SFT with long-context SFT together and only perform one long-context SFT stage, a considerable amount of training resources would be wasted.
Thus, we suggest using a short-context window, like 4K, for alignment. Then, we use a long context windows, such as 100K, to train a mix of long-context SFT data and normal SFT data. We incorporate normal SFT samples in the long-context SFT stage to preserve the model's alignment capability.

\begin{table}[t]
\centering
\fontsize{8}{13}\selectfont
\caption{\label{multi-style} Statistics of average token counts for samples and responses in the normal SFT and long-context SFT corpora after packing. To create the training data for the long-context SFT stage, we concatenate multiple SFT samples together to achieve a length of approximately 100K tokens. Regular SFT data refers to SFT data commonly used for alignment. Long-context SFT data refers to SFT data that focuses on long-context modeling, such as book QA and book summarization.
}
\begin{tabular}{l|rrr}
\toprule
\textbf{Data} & \textbf{\#Number } & \textbf{\#Tokens} & \textbf{\#Resp. Tokens} \\
\hline
Long-context SFT & 21398 & 88266 & 1959 \\
Regular SFT & 8484 & 98994 & 85582 \\

\bottomrule 
\end{tabular}
\label{tab:sft_data_distribution}
\end{table}


\subsection{Synthetic data in the long-context continued pretraining stage}


\citet{xiong2023effective} suggested joining short texts into longer ones to maintain a similar data distribution as the pretraining stage. However, in our study, we found that only joining short texts during pretraining causes the model to learn short patterns and struggle with understanding long context dependencies. We suggest a simple yet effective method to speed up this process. This method uses short texts to learn long dependencies. Specifically, we split several relatively short documents into segments and then rearrange these segments in an interleaved way to form a longer training sample.  Assume $D = \{d_1, d_2, ..., d_m\}$ is a pretraining dataset containing $m$ documents. To improve the narrative flow, we assume that the token count of all documents is n. Therefore, $d_i = \{x_i^1, x_i^2,….x_i^n\}$, where x represents words in the document. In what follows, we use the example of n=3 to illustrate our method. Previous methods for combining multiple short documents to form a longer document involve directly connecting $d_{long} = \{ x^1_1, x^1_2, x^1_n, x^2_1, x^2_3, x^2_n, x^3_1, x^3_2,x^3_n\}$. Our proposed method is to firstly split the document into multiple chunks, $d_i = \{c^i_1, c^i_2, c^i_3\}$, $d_j = \{c^j_1, c^j_2, c^j_3\}$ where c denotes the document chunk, which consists of tokens. Then we interleaved organize the chunks from multiple documents into a single one like $d_{long} =\{c^i_1, c^j_1,c^i_2,c^j_2,c^i_3,c^j_3 \}$. This process compels the model to focus on relevant long-distance information, thereby minimizing loss during the generation of the current token in the pre-training phase. Note that the word order in the document is preserved, which allows the attention mechanism to attend to the necessary information for generating current tokens. If the length exceeds the maximum length set, it is truncated to fit within the limit. If the length is shorter than the maximum length, it is padded to reach the maximum.

\subsection{Synthetic data in long-context SFT stage}


Creating lengthy and meaningful supervised data is a costly and labor-intensive task. This raises the question: can we utilize synthetic data to enhance the model's capacity to process extensive contextual information?

In this paper, we propose using program-generated synthetic data to improve the ability of Language Models to retrieve and understand long-context information. We define synthetic long-context SFT data as a task for processing tables. This is because it's simple to generate a table programmatically and control its attributes such as length and difficulty. The input table could be an IP address table, an information registration form, or even an HTML or markdown file with multiple fields and lines. To accommodate information of varying lengths, we generate tables ranging from 2k to 100K tokens. The fields and data in the table are generated using Faker~\footnote{https://github.com/joke2k/faker}. Detailed descriptions of the three main types of tasks we constructed will be provided.

\begin{itemize}
\item \textbf{Information retrieval task}: We prompt the language model to retrieve information from a table based on certain constraints, such as "Select the youngest people in this table and give me their email."

\item \textbf{CoT reasoning task}: We prompt the language model to generate a Chain-of-thought~\cite{wei2022chain} reasoning step to answer the question. As an example, when asked with "Calculating the age difference between the oldest male and the youngest female", the model first identifies the ages of the oldest male and youngest female. The final result is then computed by subtracting these two values.

\item \textbf{Global comprehension task}: This task necessitates the model to possess a thorough understanding of the structure and information presented in the table to offer precise results. For instance, when prompted to sort the table by age, the language model needs to have a comprehensive understanding of the table and then give the correctly sorted table.
\end{itemize}

\section{Experiments}
\subsection{Chunks interleaved long-Context pretraining (CIP)}
In line with previous work~\cite{roziere2023code, xiong2023effective}, we maintain the core Skywork architecture mostly unchanged for long context pretraining. We only make a required adjustment to the positional encoding, enabling the model to focus on longer context. Specifically, we lower the rotation angle (controlled by the hyperparameter "base frequency b"), which reduces the diminishing effect of RoPE for distant tokens. We changed the base from 10000 to 2600000, as advised by~\citet{liu2023scaling}, to scale a pretrained model from a 4K context window to extend it to a 200K context window. For additional information, individuals who are interested can refer to their paper.

Training large language model on long context windows can be very costly. In our case, we utilized a Skywork-3B model that was trained on 64K context windows of Redpajama~\cite{together2023redpajama} web text to evaluate the efficacy of our proposed methods. To create our final document, we combine multiple documents until they reach a length of approximately 64K. We then pad the document to 64K. If the final document exceeds 64K, we simply remove any excess tokens. It is interesting to discover the impact of the number of chunks per document on the model's performance. We heuristically divide the text input into 2 chunks, 4 chunks, and 8 chunks. The models trained with these methods are referred to as \textbf{3B-CIP-2}, \textbf{3B-CIP-4}, and \textbf{3B-CIP-8}. The baseline model is trained on a corpus where documents are directly concatenated to long ones, which we denotes as 3B-DC. The training corpus used for both the baseline model and our model is identical. Therefore, the baseline model can also be considered as a model with Chunk1.  We use the AdamW optimizer~\cite{loshchilov2017decoupled} and set the maximum learning rate to 1e-5, with a learning rate decay to 1e-6. We train each model for 500 iterations. The performance evaluation of the models after long-context continued pretraining stage focus on its ability to retrieve key information. Specifically, we employ a five-shot manner to evaluate the pretrained models, as introduced by~\citet{longchat2023}. As shown in Table ~\ref{table:3b_line_retrieval}, our methods can significantly improve the few-shot information retrieval capability of a pre-trained models. 3B-CIP-2 achieve a retrieval score of 0.514 which increase baseline model accuracy of 0.338 by about \textbf{52.07\%}. In addition, we evaluate each model using Chinese and English language modeling tasks, as well as language understanding tasks. As seen in Table~\ref{table:3b_understadning_task}, compared to the baseline method, our proposed method slightly increases the validation loss for English and Chinese language modeling. Specifically, the English web text validation loss of the 3B-CIP-2 model is 2.101, which is 0.33\% lower than the baseline model. The Chinese web text validation loss of the 3B-CIP-2 model is 2.152, which is 0.13\% lower than the baseline model. When measuring the capabilities of the English and Chinese language understanding by two popular benchmarks, C-EVAL~\cite{huang2023ceval} and MMLU~\cite{hendrycks2020measuring}, we find that our methods performed on par with the baseline model in both tasks. Due to the objective of improving the model's capacity to handle extensive contextual information, a slight decrease in normal language modeling task is anticipated. As we find that 3B-CIP-2 performs the best overall, we follow this setting to train our larger models.

 \begin{table}[t!]
	\centering
	\small
	\begin{tabular}{@{}lcccccc@{}}
		\toprule
		\textbf{Models}&
        \textbf{1K}& 
		\textbf{2K}&
        \textbf{4K}&
        \textbf{16K}&
        \textbf{64K}&
        \textbf{AVG.}\\
        \midrule
        
          3B-DC & 0.68 & 0.45 & 0.40 & 0.10 & 0.06 & 0.338 \\
          \midrule
          3B-CIP-2 & \textbf{0.93} & \textbf{0.79} & 0.58 & 0.14 & 0.13 & \textbf{0.514} \\
          3B-CIP-4 & 0.92 & 0.73 & 0.53 & 0.12 & \textbf{0.17} & 0.494 \\
          3B-CIP-8 & 0.90 & 0.68 & \textbf{0.60} & \textbf{0.19} & 0.09  & 0.492 \\
    \bottomrule
	\end{tabular}
	\caption{Results of the line retrieval task. We evaluate different models using various context windows ranging from 1K to 64K to measure its information retrieval capability.}
	
    \label{table:3b_line_retrieval}
\end{table}

 \begin{table}[t!]
	\centering
	\small
	\begin{tabular}{@{}lccccc@{}}
		\toprule
		\textbf{Models}&
        \textbf{En LM$\downarrow$}& 
	\textbf{Zh LM$\downarrow$}&
        \textbf{C-EVAL$\uparrow$}&
        \textbf{MMLU$\uparrow$}\\
        \midrule
        
          3B-DC & \textbf{2.094} & \textbf{2.149} & 28.47 & 28.09 \\
          \midrule
          3B-CIP-2 & 2.101 & 2.152 & \textbf{28.57} & 28.10  \\
          3B-CIP-4 & 2.100 & 2.156 & 28.29 & \textbf{28.12}  \\
          3B-CIP-8 & 2.105 & 2.157 & 28.20 & 27.95   \\
    \bottomrule
	\end{tabular}
	\caption{Results of the language model and language understanding tasks. "EN LM" denotes English web text validation loss, "ZH LM" denotes Chinese web text validation loss.}
	
    \label{table:3b_understadning_task}
\end{table} 

\subsection{Synthetic Long-context SFT (SynL)}
Following an long-context pretraining stage, we performs an SFT stage to align the model, with the context window set to 4k. Please note that during the long-context continued pretraining stage, the SFT stage, and the long-context SFT stage, the model structure and base in RoPE remain constant, set at 260000.

In the long-context SFT stage, the data source can be divided into three categories. The first category includes normal SFT data, maintained to preserve alignment capability. The second category involves naturally collected long-context SFT data, such as book QA and book summarization. We refer to this data as true long-context SFT data. The third category pertains to our synthetic long-context SFT data. Based on our preliminary experiments, we found that maintaining the ratio of normal SFT data at 20\% yielded the best performance. Therefore, in the following tests, we will keep the ratio of normal SFT data unchanged.

We evaluate the long-context capability of each model on InfiniteBench~\cite{zhang2023infinitebench}, which is designed to evaluate the capabilities of language models to process, understand, and reason over super long contexts (100k+ tokens). This benchmark involves tasks related to long-context retrieval and understanding in English, Chinese, Code, and Math.

\subsubsection{Baselines}

We compared our methods with several proprietary model APIs, such as GPT4, Claude, Moonshot, and several open-sourced baseline models as follows:

\begin{itemize}
\item \textbf{Yarn-Mistral-7B-128K}~\cite{peng2023yarn}: This model is  a long-context model presented by NousResearch\footnote{https://nousresearch.com/}. The model is based on Mistral~\cite{jiang2023mistral} and using Yarn~\cite{peng2023yarn} to extend the context windows to 128K. 

\item \textbf{Yarn-Llama2-13B-128K}~\cite{peng2023yarn}: This model is based on Llama2-13B~\cite{touvron2023llama2} and using Yarn to extend the context windows to 128K. 

\item \textbf{LongSkywork-13B-T}: This model is a variant of LongSkywork-13B which use 20\% normal SFT data and 80\% true long-context SFT data collected from LongAlpaca~\cite{long-alpaca} and LongCollection~\footnote{https://huggingface.co/datasets/togethercomputer/Long-Data-Collections}.

\item \textbf{LongSkywork-13B-S}: This model is a variant of LongSkywork-13B, which uses 20\% normal SFT data and 80\% long-context synthetic SFT data.

\item \textbf{LongSkywork-13B-S\&T}: This model is a variant of LongSkywork-13B, which uses 20\% normal SFT data, 40\% true long-context SFT data, and 40\% synthetic long-context SFT data.
\end{itemize}

\begin{table*}[ht!]
	\centering
	\small
	\setlength{\tabcolsep}{4pt}
	\scalebox{0.90}{\begin{tabular}{lcccccccccccc}
	\toprule
	\textbf{Models} &
        \textbf{Re.PKey} & 
	\textbf{Re.Num}&
        \textbf{Re.KV} &
        \textbf{En.Dia} &
        \textbf{En.Sum} &
        \textbf{En.MC} &
        \textbf{En.QA} &
        \textbf{Zh.QA} &
        \textbf{Math.Find} &
        \textbf{C.Debug} & 
        \textbf{Re.Avg} & 
        \textbf{Avg}
        \\ \midrule
          GPT-4* & 100.0\% & 100.0\% & 89.0\% & 22.2\% & 14.7\% & 67.3\% & 8.5\% & 24.3\% & 60.0\% & 39.6\% & 96.3\% &52.6\%\\
          Claude2.1*  & 97.8\% & 98.1\% & 65.4\% & 46.5\% & 14.5\% & 62.9\% & 12.0\% & 10.5\% & 32.3\% & <5\% & 87.1\% & 44.5\%\\
          Moonshot*  & 98.1\% & 95.4\% & 34.2\% & 16.5\% & 17.9\% & 72.5\% & 11.5\% & 17.3\% & 12.6\% & 18.0\% & 87.1\% & 39.4\%\\
          Y-Mistral-7B*  & 92.7\% & 56.6\% & <5\% & 9.6\% & 9.1\% & 28.0\% & 7.5\% & 14.4\% & 17.1\% & <5\% & 51.4\% & 24.5\%\\
          Y-Llama2-13B  & 44.0\% & <5\% & <5\% & 5.0\% & 12.0\% & 32.0\% & 8.7\% & 6.4\% & 11.0\% & <5\% & 18.0\% & 13.4\%\\
          \midrule
          LongS-13B-T  & 45.0\% & 98.0\% & <5\% & 7.0\% & 16.4\% & 37.0\% & 6.2\% & 11.6\% & <5\% & 17.0\% & 49.3\% & 24.7\%\\
          LongS-13B-S  & 100.0\% & 95.0\% & 87.0\% & 6.0\% & 8.5\% & 33.0\% & 5.6\% & 15.1\% & <5\% & 26.0\% & 94.0\% & 37.8\%\\
          LongS-13B-S\&T  & 86.0\% & 85.0\% & 69.0\% & 12.0\% & 14.9\% & 40.0\% & 6.8\% & 18.1\% & 30.0\% & 28.0\% & 80.0\% & 39.0\%\\
          LongS-13B  & 100.0\% & 99.0\% & 98.0\% & 10.0\% & 14.0\% & 46.0\% & 31.2\% & 28.0\% & 29.0\% & 20.0\% & 99.0\% & 47.5\%\\
    \bottomrule
	\end{tabular}}
	\caption{Results on InfiniteBench: we removed the tasks of MATH.Calc and Code.Run, as most models showed less than 5\% accuracy in these tasks. Performing long context inference requires a significant amount of resources and is time-consuming. Therefore, when evaluating tasks with more than 100 numbers, we only consider the performance of the first 100 samples. * denotes we use the result test by official repertory of InfiniteBench. }
    \label{table:infinite_result}
\end{table*}

The results are outlined in Table~\ref{table:infinite_result}. Our proposed synthetic data is highly efficient. LongSkywork-13B-S achieves an average score of 37.8\%, outperforming LongSkywork-13B-T by a large margin, which scores 24.7\%. We find synthetic data can significantly enhance the long-context retrieval capability of an LLM, raising it from 49.3\% to 94.0\% on average across three retrieval-based tasks. In addition to the retrieval-based tasks, synthetic data can also improve the Chinese QA task from 11.6\% to 15.1\%, demonstrating that our methods can enhance the long-context understanding of an LLM. However, we observed a decrease in the EN.sum task from 16.4\% to 8.5\% because there is no summary-related data in the synthetic data. This leaves the construction of summary-related data as future work. Upon merging synthetic data with actual long-SFT data, we observe an uplift in the average score, albeit with a minor decrease in retrieval-based tasks. We believe this is due to an imprecise data mixing ratio. Our refined model, LongSkywork-13B, enhances this data mixing and incorporates selected annotator data. The end result is a performance of 47.5\%, outperforming both Moonshot and Claude2.1. Note that compared to other proprietary models, our model may have fewer parameters, specifically just 13 billion parameters.

To evaluate the effectiveness of LongSkywork in realistic-usage scenarios, we create a benchmark based on actual use cases from the Tiangong AI Reading website~\footnote{https://work.tiangong.cn/chatdoc/d/doc/index}. We collect 856 questions of various types, including questions about details in a document, document summarization, and questions that cannot be answered based on the document. We then ask the annotators to provide a reference answer for each question. We used GPT-4 for evaluation, comparing the reference answers to the answers provided by the models. The results are displayed in Table~\ref{table:poc_test}. Our LongSkywork-13B model attains an overall score of 69.24, which is slightly lower than the state-of-the-art long-context model Claude2.1, scoring 72.27. This indicates that our method demonstrates effective performance when it comes to real-world scenario questions.

 \begin{table}[t!]
	\centering
	\small
	\begin{tabular}{@{}lcccc@{}}
		\toprule
		\textbf{Models}&
        \textbf{Wrong}& 
	\textbf{Part.Right}&
        \textbf{Right} & 
        \textbf{Overall} \\
        \midrule
        
          Claude2.1 & 4.67\% & 34.58\% & 60.75\% & 72.27 \\
          \midrule
          LongS-3B & 10.51\% & 43.22\% & 46.03\%  & 60.44 \\
          LongS-13B & 7.13\% & 35.28\% & 57.48\%  & 69.24 \\
    \bottomrule
	\end{tabular}
	\caption{Results on a realistic application scenario benchmark. Part.Right denotes partially correct. The score are judged by GPT-4. Overall score are calculated by (3 * Right + Part.Right ) / 3. LongS stands for our LongSkywork model. }
	
    \label{table:poc_test}
\end{table} 

\subsubsection{Limitations and future work}
In this section, we discuss some limitations of our proposed methods for future research. We identify two issues with LongSkywork as shown in Table~\ref{table:case_study}. First, it struggles to effectively handle non-answerable questions, likely due to a limited number of non-answerable training samples. Second, it has difficulty with long-context complex reasoning and calculations, possibly due to the challenge of deriving correct reasoning steps from a 13B parameter base model. In the future, we will strive to enhance the two aforementioned aspects of LongSkywork.

\begin{table}[!t]
\footnotesize
\small
\centering
\scalebox{0.92}{ \begin{tabular}{|p{1.00\linewidth}|}
    \hline
    \textbf{Input 1}: \textsl{In which countries and regions is the market size larger for games independently developed in our country in 2022?}\\
      \textbf{Reference Answer}:  \textsl{The report did not specify any particular country or region, only mentioning that developed markets like the United States and emerging markets such as the Middle East, Africa, and Latin America have great potential.}    \\  

      \textbf{LongSkywork-13B}: \textsl{In 2022, the games independently developed by our country have the largest market share in the United States, accounting for 32.31\%; followed by Japan, accounting for 17.12\%; South Korea accounts for 6.97\%.}\\ 

    \hline
    \textbf{Input 2}: \textsl{How many people died in the 9/11 events, excluding those in the New York World Trade Center and the Pentagon in Washington?}\\
      \textbf{Reference Answer}:  \textsl{The "9·11" incident resulted in a total of 3126 deaths. Of these, 2550 people died in the World Trade Center in New York; 343 people died on American Airlines Flight AA11 and United Airlines Flight UA175; 125 people died in the Pentagon in Washington; 64 people died on American Airlines Flight AA77; 44 people died on United Airlines Flight UA93 in Pennsylvania. Therefore, excluding the World Trade Center in New York and the Pentagon in Washington, a total of 3126-2250-125=751 people died.}    \\  

      \textbf{LongSkywork-13B}: \textsl{About 343 people died in the "9·11" event, excluding the New York World Trade Center and the Pentagon in Washington.}\\ 
\hline

    \end{tabular}}
    \caption{Case studies on two limitations of LongSkywork-13B. }
    \label{table:case_study}
\end{table}

\section{Conclusion}
In this paper, we present the training recipe of LongSkywork and propose two methods for constructing synthetic data to enhance the long context retrieval and understanding capabilities of LLMs. The first method, referred to as CIP, reorganizes chunks of training documents to create a long training sample. This approach helps LLMs learn long-term dependencies more effectively. The second method, referred to as SynL, generates a variety of long-context question pairs, which can enhance reasoning and tackle the issue of data scarcity common in long-context SFT. We conduct extensive experiments to evaluate the effectiveness of the proposed methods. In InfiniteBench, LongSkywork-13B achieves an average score of 47.5, matching or exceeding the performance of top-tier proprietary LLM APIs. In a real-world usage evaluation, our LongSkywork-13B model performance only slightly decreases compared to Claud2.1. We also pinpoint two limitations of LongSkywork: handling non-answerable questions and dealing with complex, long reasoning problems. We will prioritize these topics in our future work.

\bibliography{anthology,custom}
\bibliographystyle{acl_natbib}

\appendix

\end{document}